\documentclass[sn-mathphys-num]{sn-jnl}

\usepackage{graphicx}%
\usepackage{multirow}%
\usepackage{amsmath,amssymb,amsfonts}%
\usepackage{amsthm}%
\usepackage{mathrsfs}%
\usepackage[title]{appendix}%
\usepackage{xcolor}%
\usepackage{textcomp}%
\usepackage{manyfoot}%
\usepackage{booktabs}%
\usepackage{algorithm}%
\usepackage{algorithmicx}%
\usepackage{algpseudocode}%
\usepackage{listings}%
\usepackage{subcaption}
\usepackage{adjustbox}
\usepackage{multirow}
\usepackage{makecell}
\usepackage{rotating}

\theoremstyle{thmstyleone}%
\theoremstyle{thmstyletwo}%

\theoremstyle{thmstylethree}%

\begin{document}

\title[]{Modality-missing RGBT Tracking: Invertible Prompt Learning and High-quality Benchmarks}


\author[1]{\fnm{Andong} \sur{Lu}}\email{adlu\_ah@foxmail.com}

\author[2]{\fnm{Jiacong} \sur{Zhao}}\email{JiacongZhao2022@163.com}

\author*[2]{\fnm{Chenglong} \sur{Li}}\email{lcl1314@foxmail.com}
\author[1]{\fnm{Jin} \sur{Tang}}\email{tangjin@ahu.edu.cn}
\author[1]{\fnm{Bin} \sur{Luo}}\email{luobin@ahu.edu.cn}

\affil*[1]{\orgdiv{School of Computer Science and Technology}, \orgname{Anhui University}, \orgaddress{\street{Street}, \city{Hefei}, \postcode{230601}, \state{Anhui}, \country{China}}}

\affil*[2]{\orgdiv{School of Artificial Intelligence,}, \orgname{Anhui University}, \orgaddress{\street{Street}, \city{Hefei}, \postcode{230601}, \state{Anhui}, \country{China}}}


\abstract{
Current RGBT tracking research relies on the complete multi-modal input, but modal information might miss due to some factors such as thermal sensor self-calibration and data transmission error, called modality-missing challenge in this work. To address this challenge, we propose a novel invertible prompt learning approach, which integrates the content-preserving prompts into a well-trained tracking model to adapt to various modality-missing scenarios, for robust RGBT tracking.
Given one modality-missing scenario, we propose to utilize the available modality to generate the prompt of the missing modality to adapt to RGBT tracking model. However, the cross-modality gap between available and missing modalities usually causes semantic distortion and information loss in prompt generation. To handle this issue, we design the invertible prompter by incorporating the full reconstruction of the input available modality from the generated prompt.
To provide a comprehensive evaluation platform, we construct several high-quality benchmark datasets, in which various modality-missing scenarios are considered to simulate real-world challenges.
Extensive experiments on three modality-missing benchmark datasets show that our method achieves significant performance improvements compared with state-of-the-art methods. We have released the code and simulation datasets at: \href{https://github.com/Alexadlu/Modality-missing-RGBT-Tracking.git}{https://github.com/Alexadlu/Modality-missing-RGBT-Tracking.git}.
}




\maketitle

\section{Introduction}\label{sec1}

\begin{figure}[t]
     \centering
     \begin{tabular}[b]{cc}
     
     \includegraphics[width=0.65\textwidth ]{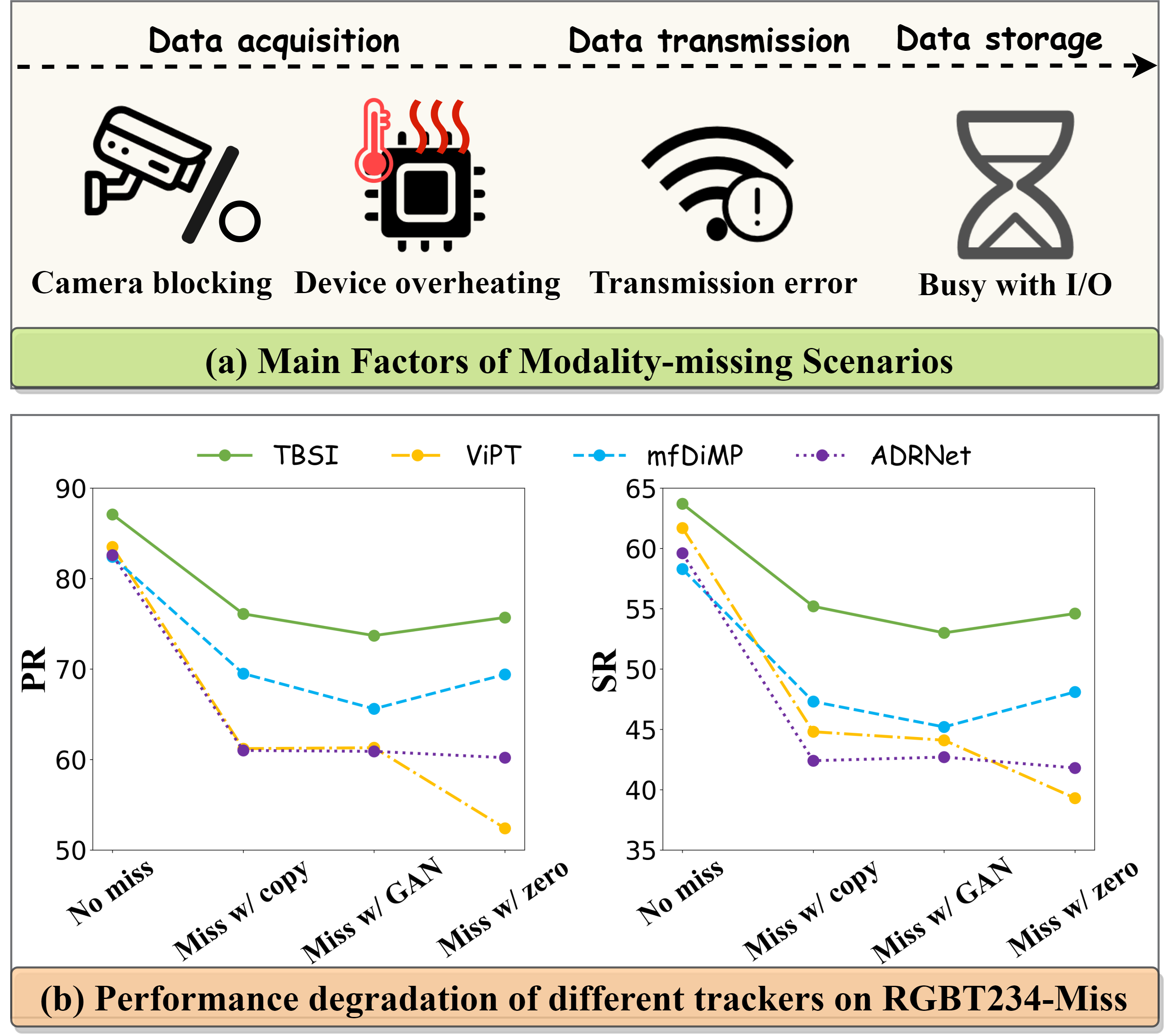}
     \end{tabular}
     \caption{Illustration of modality-missing factors and evaluation results of state-of-the-art methods on the RGBT234-Miss dataset.}
     \label{figure:performance_de}
\end{figure}

In recent years, RGBT tracking attracts widespread attention in the field of computer vision~\cite{2021MANet++, cui2022visual, zhang2023efficient, Zhang_CVPR22_VTUAV, TBSI, ViPT, APFNet2022, DMCNet2022, 2020CMPP}. Thanks to powerful neural networks such as Transformer~\cite{ViT_network}, and large-scale RGBT datasets such as LasHeR~\cite{li2021lasher} and VTUAV~\cite{Zhang_CVPR22_VTUAV}, significant progress has been made in this field. Existing methods achieve impressive performance in modality-complete scenario by effectively utilizing information from different modalities. However, a critical modality-missing challenge in real-world scenarios is overlooked, as shown in Figure~\ref{figure:performance_de} (a). There are many factors causing the modality-missing challenge, such as thermal sensor self-calibration and data transmission errors. Existing RGBT trackers rely on the complete RGB and thermal data, and are thus hard to handle this challenge. 

We evaluate several state-of-the-art RGBT trackers on modality-missing dataset. To enable these trackers to be executed on modality-missing data, we use three widely used strategies to compensate for the missing data, including “Zero”, “Copy”, “GAN”. In specific, the “Zero” strategy is to create a zero-valued matrix as the missing input, “Copy” is to use the available modality as the missing input, and “GAN” is to generate the missing input from the available modality using the generative adversarial network (GAN)~\cite{pix2pix_gan}. The evaluation results of different trackers using different compensation strategies are presented in Figure~\ref{figure:performance_de} (b). From the results we can see that, although compensation strategies are employed, these trackers still exhibit significant performance degradation in suffering from the modality-missing challenge. Therefore, how to perform robust RGBT tracking under the modality-missing challenge, called modality-missing RGBT tracking in this work, is a vital research topic that needs to be solved urgently. 

To handle the modality-missing challenge, this work presents a novel invertible prompt learning approach, which integrates the content-preserving prompts into a well-trained tracking model to adapt to various modality-missing scenarios, for modality-missing RGBT tracking. Data compensation strategies such as “Zero”, “Copy” and “GAN” can not meet the requirement of tracking models in modality-missing scenarios, and thus greatly degrade tracking performance. To this end, we introduce the idea of prompt learning to maintain tracking performance in modality-missing scenarios. In particular, we design a prompt generation model that utilizes available modality information in the current frame to generate the prompts for the missing modality. Since all modalities are accessible during the training phase, we can utilize missing modality features to guide the learning of prompts, and thus the generated prompts can well meet the requirements of the tracking model in missing scenarios.

However, the cross-modality gap between available and missing modalities usually causes semantic distortion~\cite{kingma2018glow, Zhou2023homeomorphism} and information loss~\cite{gan2017semantic, cao2018partial} in cross-modality generation. 
To address these issues, we design an invertible prompter that incorporates a full reconstruction constraint from the generated prompt to the input available modality. 
We observe that the issues of semantic distortion and information loss could cause irreversible cross-modal generation. Based on this observation, we introduce an inverse reconstruction scheme in the prompter generation to improve the quality of generated prompts.
These designs ensure that our prompt generation process can preserve content and avoid semantic distortion. In addition, we also introduce a task alignment loss to encourage the invertible prompter to learn more discriminative prompt information, thereby reducing the gap between the invertible prompts and the downstream tracking task. During the testing phase, our invertible prompter is activated only when modality-missing occurs, thereby avoiding any unnecessary computational burden in modality-complete scenarios.

Although many high-quality datasets are available in the RGBT tracking field, there still lacks a modality-missing dataset. However, as discussed above, the factors causing modality-missing scenarios in real scenes are complex and diverse, making it difficult to capture all kinds of modality-missing data. To handle this issue, we propose a high-quality data simulation approach to construct several high-quality benchmark datasets, in which various modality-missing scenarios are considered to simulate real-world challenge. In particular, we design a hierarchical combination scheme based on three typical missing patterns to simulate possible modality-missing scenarios. Our data simulation approach is generic and can be applied to different modality-complete datasets. In this work, we select three representative RGBT tracking datasets, namely RGBT234~\cite{li2019rgb234}, LasHeR~\cite{li2021lasher}, and VTUAV~\cite{Zhang_CVPR22_VTUAV}, to construct three modality-missing RGBT tracking benchmarks, corresponding RGBT234-Miss, LasHeR245-Miss and VTUAV176-Miss. In these datasets, we conduct a comprehensive evaluation of state-of-the-art RGBT trackers in recent years, and these results will lay the research foundation for this field.

In summary, this paper investigates the issue of missing modalities in RGBT tracking and proposes a new solution and several high-quality benchmark datasets for modality-missing RGBT tracking. The main contributions are as follows:

\begin{itemize}
\item We systematically study the modality-missing problem in RGBT tracking and introduce a prompt learning framework to solve the modality-missing problem in RGBT tracking for the first time.

\item We propose a novel invertible prompt learning approach, which effectively addresses the issues of semantic distortion and information loss in cross-modality prompt generation. Furthermore, a task alignment loss is introduced to narrow the gap between generated prompts and downstream tracking task.

\item We construct several high-quality benchmark datasets to provide a comprehensive evaluation platform for modality-missing RGBT tracking. A hierarchical combination of missing patterns is designed to simulate possible modality-missing scenarios in the real world.

\item Extensive experiments on both public and created datasets suggest that our method achieves outstanding performance against state-of-the-art RGBT tracking methods in both modality-missing and modality-complete scenarios.
\end{itemize}

\section{Related Work}

\subsection{RGBT Tracking Methods}
Due to the complementarity of thermal infrared and visible information, RGBT tracking becomes a promising research topic in the field of computer vision and obtains widespread attention in recent years. With the continuous emergence of RGBT datasets~\cite{li2019rgb234, Li17rgbt210,li2021lasher,Zhang_CVPR22_VTUAV} and the vigorous development of deep neural networks~\cite{deng2009imagenet, ViT_network}, the development of this field is further promoted. Thanks to the powerful representation capabilities of neural networks, deep trackers dominate the field. They can be divided into three categories based on different tracking frameworks. Early trackers, based on the MDNet series~\cite{li2019manet,2021MANet++,2020CAT, APFNet2022, 2020CMPP}, can learn from the limited-scale RGBT tracking data and achieve impressive performance. Another Siamese network-based trackers~\cite{zhang2019multi,zhang2021siamcda} achieves a balance between performance and efficiency based on the generated large-scale RGBT dataset. With the release of large-scale RGBT tracking 
datasets~\cite{li2021lasher,Zhang_CVPR22_VTUAV}, current Transformer-based trackers achieve great success in this field~\cite{ViPT, ProRGBTTrack, TBSI}. However, existing methods and benchmarks are still developed for modality-complete scenarios, while ignores the modality-missing problem in real-world scenarios. Therefore, this work investigates the modality-missing RGBT tracking task, and designs an effective modality-missing tracking algorithm and creates several high-quality modality-missing RGBT tracking benchmarks.

\subsection{Modality-missing Learning Methods}

Currently, many research fields are focusing on the problem of multi-modal learning in modality-missing scenarios. To address this issue, various methods have been proposed. For instance, Ma~\emph{et al.}~\cite{ma2021smil} introduces a Bayesian meta-learning-based approach for reconstructing latent features of missing modalities. Zhao~\emph{et al.}~\cite{zhao2021missing} presents a missing modality imagination network for prediction using available modalities. Ma~\emph{et al.}~\cite{ma2022multimodal} explore the robustness of multi-modality Transformer in missing scenarios and enhance it through multitask optimization. 

In multi-view learning, several studies~\cite{yin2017unified, zhang2020deep, xu2022deep, lin2021completer} tackle the issue of missing views by developing effective representation learning approaches. For example, Yin~\emph{et al.}~\cite{yin2017unified} create a unified subspace by leveraging incomplete or unlabeled multi-view data. Zhang~\emph{et al.}~\cite{zhang2020deep} propose an adversarial strategy to generate representations of missing views. 
In medical image analysis, some studies~\cite{hu2020knowledge, wang2021acn, wang2023multi} focus on distilling the knowledge of full-modality models into specific models to handle modality-missing scenes. However, these methods often introduce highly complex generative models or a combination of multiple specific models to handle different missing scenarios, which poses challenges in practical deployment. Moreover, Yi~\emph{et al.}~\cite{lee2023multimodal} introduce the concept of prompting to handle missing text or images of visual recognition. However, their method provides the same prompts for missing instances, which limits its application in tracking. In this paper, we design an innovative invertible prompt learning method, which generates instance-aware prompts to handle the modality-missing challenge in RGBT tracking.

\section{Methodology}
\label{sec:me}

\subsection{RGBT Tracking Baseline Model}
We first investigate how to design a more robust RGBT tracking model under modality-missing scenarios through multiple design choices. Existing RGBT tracking frameworks usually adopt an architecture of a feature extraction network followed by a fusion unit. To this end, we conduct the following verification, including the design of the feature extraction network and the selection of the fusion unit, to more deeply analyze the impact on the RGBT tracking model in the modality-missing scenario.

\noindent\textbf{Architecture of Feature Extractor.} 

\begin{figure}[h]
     \centering
     \begin{tabular}[b]{cc}
     \includegraphics[width=1\textwidth]{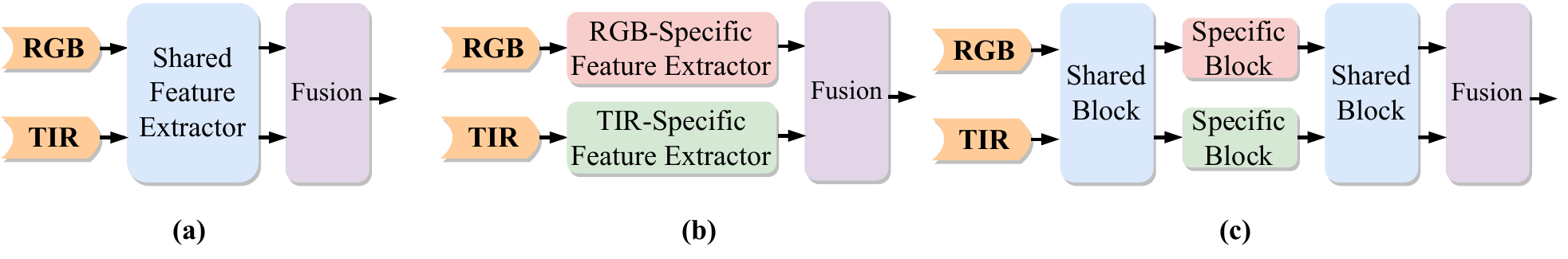}
     \end{tabular}
     \caption{Three feature extractors in RGBT tracking models.}
     \label{figure:feature_extractor_compare}
\end{figure}

As shown in Figure~\ref{figure:feature_extractor_compare}, the mainstream feature extractor of RGBT tracking can be categorized into three types: (a) shared feature extractor, (b) specific feature extractor, and (c) shared-specific feature extractor. We seek the optimal feature extractor in modality-complete and modality-missing scenarios, in which the fusion unit is set to simple summation by default.
To ensure a fair comparison, we evaluate these three feature extraction networks within the same tracking framework. Specifically, we follow recent powerful Transformer-based tracking approach~\cite{ostrack} to extend ViT~\cite{ViT_network} as a multi-modality extractor of our baseline tracker. Table~\ref{tab_fe_exp} compares the performance of three feature extractors on RGBT234 and RGBT234-miss datasets (See details in next section), and it can be observed that the shared-specific design demonstrates advantages in both modality-complete and modality-missing scenarios. 

\begin{table*}[h]
\centering
\caption{Evaluation of different extractors on RGBT234 and RGBT234-Miss datasets.}
\renewcommand\arraystretch{1.4}
\resizebox{1\columnwidth}{!}{
\label{tab_fe_exp}
\begin{tabular*}{\textwidth}{@{\extracolsep\fill}cl|cccc}
\toprule
\multirow{2}{*}{Extractor} & \multirow{2}{*}{Fusion} & \multicolumn{2}{c}{RGBT234-Miss} & \multicolumn{2}{c}{RGBT234 } \\ 

                                &                         & PR              & SR              & PR           & SR              \\ \hline
(a)                             & \multirow{3}{*}{Sum}  &   74.8              &  54.4               &    86.4          &   64.5         \\
(b)                             &                         &   74.7              &  54.0               &    86.6          & 64.3          \\
(c)                             &                         & \textcolor{blue}{75.4}               & \textcolor{blue}{54.9}               &  \textcolor{blue}{88.0}           & \textcolor{blue}{65.0} \\\bottomrule

\end{tabular*}}

\end{table*}

Hence, we choose the shared-specific architecture as the feature extractor of our model. The details of this architecture are as follows. For given RGB modality and thermal infrared (TIR) modality, the search and template frames are first partitioned into patches with the size of $p\times p$ using a learnable patch embedding layer, and flattened to obtain four token sequences: $S_{rgb}^p$, $S_{tir}^p$, $T_{rgb}^p$, and $T_{tir}^p$. Following~\cite{ostrack}, we add a learnable position embedding with these token features to provide positional prior information. Note that we share the position embedding between the two modalities since the original frames are highly aligned. Afterward, we concatenate the RGB and TIR tokens, denoted as $I_{rgb} = [S_{rgb}^p, T_{rgb}^p]$ and $I_{tir} = [S_{tir}^p, T_{tir}^p]$, to feed into a set of shared Transformer blocks for extracting modality-shared features, respectively.

In addition, existing works~\cite{2021MANet++,li2019manet} already proved that modality-specific information is crucial for RGBT tracking. Hence, we employ modality-specific Transformer blocks~\cite{ViT_network} to model modality-specific features for each modality. We also introduce two learnable tokens $G_{rgb}$ and $G_{tir}$ for each modality to model modality-specific global features. Thus, the inputs of two modality-specific blocks are $I_{rgb}^* = [G_{rgb}, I_{rgb}]$ and $I_{tir}^*= [G_{tir}, I_{tir}]$. Note that $G_{rgb}$ and $G_{tir}$ are only used in modality-specific Transformer blocks, while removed in modality-shared Transformer blocks. Finally, we alternately combine modality-shared and modality-specific Transformer blocks to build the backbone model.

\noindent\textbf{Fusion Unit.}
In Table~\ref{tab_fusion_exp}, we study the inference performance changes of different fusion schemes in modality-missing and modality-complete scenarios. Specifically, we employ three simple yet effective fusion strategies that are widely used in existing RGBT tracking models, including summation, concatenation, and Transformer, as follows:

\begin{itemize}
\item \textbf{Sum:} Two modality features extracted from the backbone network $\Theta$ are summed up: $\underset{m}{\sum}\Theta (I_m) $.
\item \textbf{Concat:} Two modality features extracted from the backbone network are channel-wise concatenated ($\complement$): $\underset{m}{\complement} \Theta (I_m)$. 
\item \textbf{Transformer:} Two modality features are first token-wise concatenated and fed into two Transformer blocks with position embedding. Then the output features are divided into two features from the token dimension,  and then channel-wise concatenated.
\end{itemize}

\begin{table*}[h]
\centering
\caption{Evaluation of different fusion units on RGBT234 and RGBT234-Miss datasets.}
\renewcommand\arraystretch{1.4}
\resizebox{1\columnwidth}{!}{
\label{tab_fusion_exp}
\begin{tabular*}{\textwidth}{@{\extracolsep\fill}c|cccccc}
\hline
\multirow{2}{*}{Fusion} & \multicolumn{2}{c}{RGBT234-Miss} & \multicolumn{2}{c}{RGBT234} & \multicolumn{2}{c}{$\Delta\downarrow$} \\ 
                                & PR              & SR              & PR           & SR       & PR           & SR     \\ \hline
Sum                             &  75.4            & 54.9             &  88.0       & 65.0       &-12.6  & -10.1       \\
Concat                          & \textcolor{blue}{77.2}                & \textcolor{blue}{56.0}                &  \textcolor{blue}{88.3}             &\textcolor{blue}{65.7}      & \textcolor{blue}{-11.1}  & \textcolor{blue}{-9.7}      \\
Transformer                    &   74.8              &  54.4               &  88.0            &  64.8      & -13.2  &  -10.4 \\ \hline
\end{tabular*}}
\end{table*}

From the above results, the concatenation method achieves the best performance in two scenarios, while Transformer fusion presents the worst performance in modality-missing scenarios. It is also revealed in~\cite{ma2022multimodal} that Transformers without specially designed ones are less robust in modality-missing scenarios. Based on the above observations, we build a robust RGBT tracking baseline model, which comprises a shared-specific Transformer network as a backbone, and a simple and effective concatenation strategy to perform fusion.

\subsection{Invertible Prompt Model}

\begin{figure*}[ht]
     \centering
     \begin{tabular}[b]{cc}
      \includegraphics[width=1\textwidth]{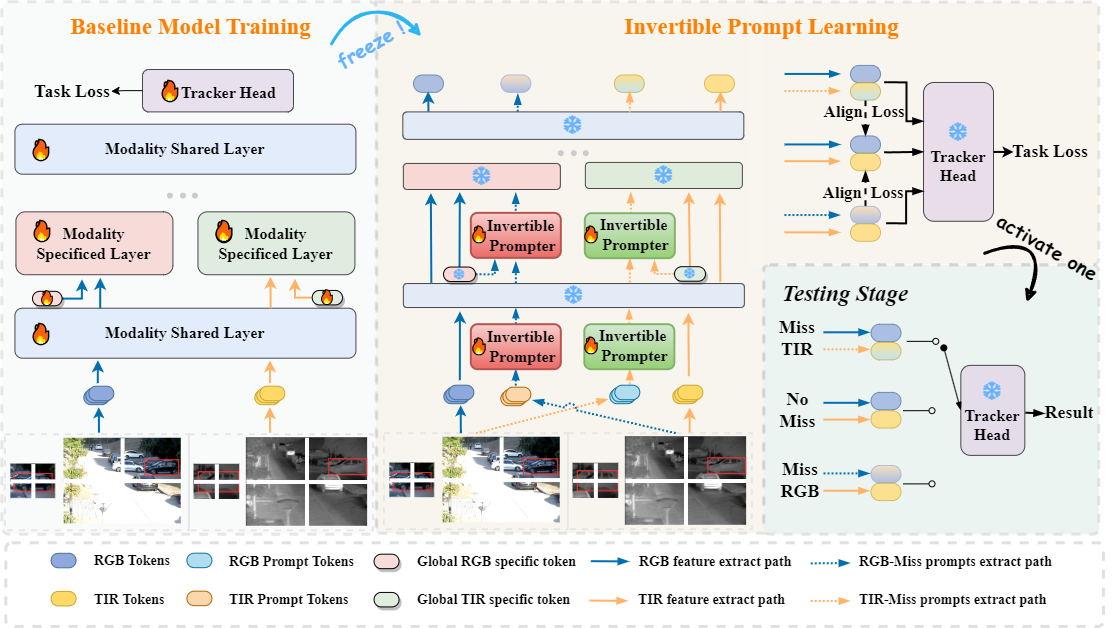}
     \end{tabular}
     \caption{Modality-missing RGBT tracking framework with the invertible prompt learning.}
     \label{figure:network_overview}
\end{figure*}

\begin{figure}[ht]
     \centering
     \begin{tabular}[b]{cc}
     \includegraphics[width=1\textwidth]{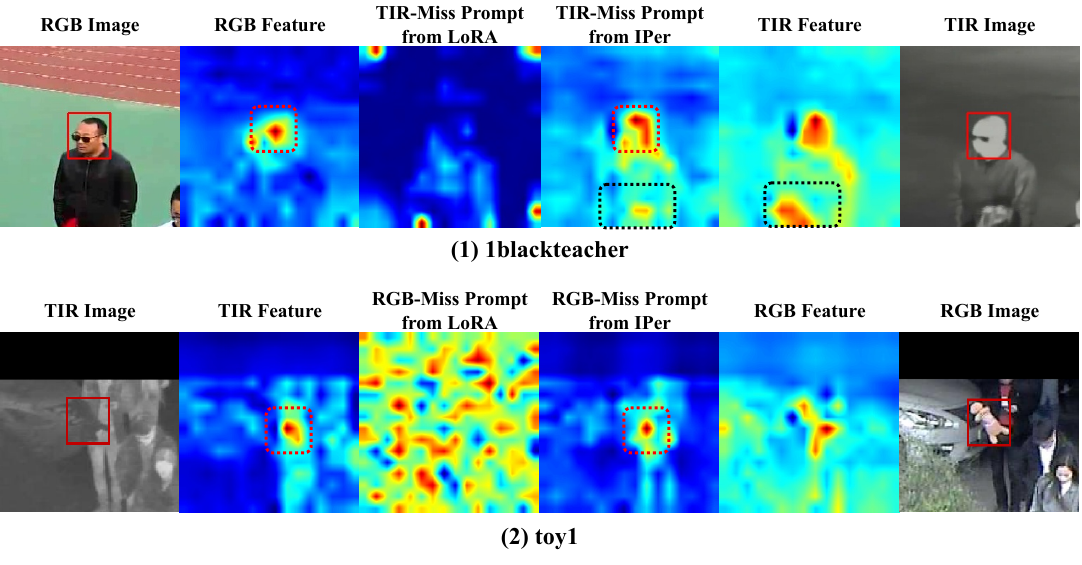}
     \end{tabular}
     \caption{Visualization of feature maps between the proposed Invertible Prompter (IPer) and popular adapter LoRA~\cite{hu2021lora}.}
     \label{figure:visual_feature}
\end{figure}

To solve the modality-missing issue, we propose an Invertible Prompt Learning (IPL) approach, as shown in Figure~\ref{figure:network_overview}. Recently, a few works begin to explore introducing some learnable parameters to the frozen well-trained model to learn effective prompts. By merely fine-tuning a small number of parameters for prompts, they achieve surprising performance improvements. 
However, the challenge of prompt generation lies in not only narrowing the gap between upstream and downstream tasks~\cite{hu2021lora, adapter_2023} and fusing different modality information~\cite{ViPT}, but also generating content-preserving cross-modal prompt.
Due to the modality gap, existing adapters~\cite{hu2021lora,ViPT} are hard to handle semantic distortion and information loss problem in the cross-modality prompt generation, as shown in the third column of Figure~\ref{figure:visual_feature}. 

To deal with this issue, we design an invertible prompter to achieve content-preserving cross-modality generation. The invertible prompter can generate high-quality prompt features for the missing modality from the available modality. The prompt features, on the one hand, obtain strong discriminative information than the available modality, and on the other hand, suppress the noise in the missing modality, as shown in the red and black rectangular boxes in the Figure~\ref{figure:visual_feature} (1). 
In addition, the invertible prompter can maintain stable generation quality and provide discriminative information, even if the modality-missing ground-truth features used for supervise prompt generation are with poor quality (as depicted in the fourth column of Figure~\ref{figure:visual_feature} (2)). In contrast, existing adapter LoRA~\cite{hu2021lora} fails to perform under similar conditions.

Formally, we provide the token sequences $\mathcal{F}_{am}$ and $\mathcal{F}_{mm}$ for the available and missing modalities, respectively. We also include a frozen RGBT tracker encoder $\mathit{E}$, which comprises $\mathit{N}$ Transformer blocks \{$\mathit{E}^n$\}$^N_{n=1}$. The designed invertible prompter aims to learn to generate the prompt features aligned with missing modality from the available modality, which can be written as: 
\begin{equation} \label{eq1}
\begin{aligned}
\mathcal{P}_{mm}^n &= \mathcal{IP}^n(\mathit{E}^n(\mathcal{F}_{am}^{n-1}))\quad n = 1, 2, ..., N , 
\end{aligned}
\end{equation}
where $\mathcal{P}_{mm}^n$ represents the prompt features for the missing modality in the $n$-th layer. 
By embedding prompters in each layer of the feature extractor, the difficulty of cross-mode generation of each prompter can be effectively reduced, which has been successfully proved in existing methods~\cite{ViPT,hu2021lora}.
Furthermore, we integrate the task alignment losses at the predictive encoding levels to enhance the discriminability of prompts.

\begin{figure}[ht]
     \centering
     \begin{tabular}[b]{cc}
     \includegraphics[width=1\textwidth]{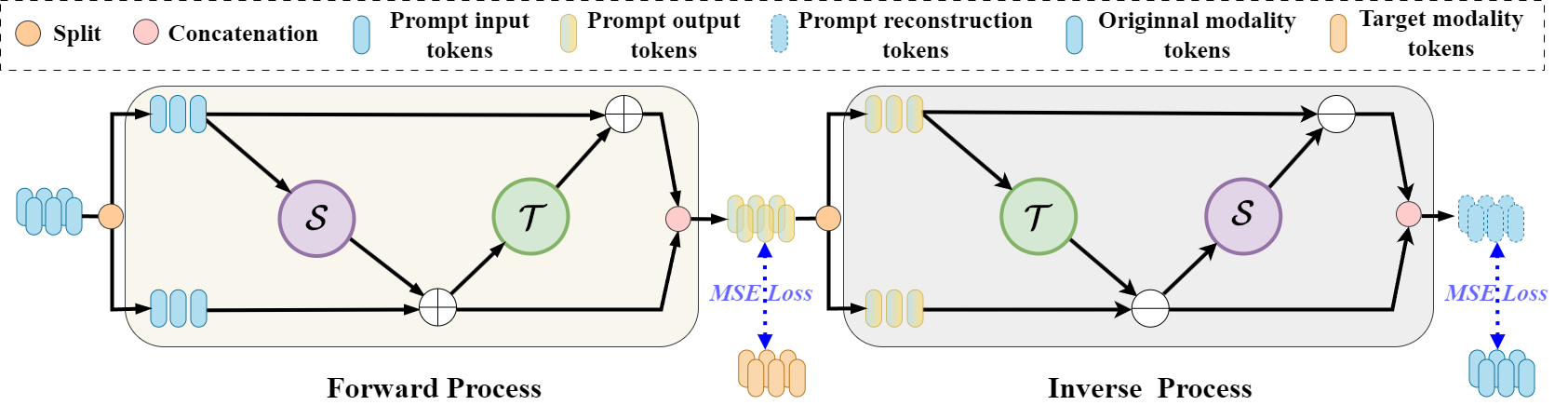}
     \end{tabular}
     \caption{Illustration of Invertible Prompter (IPer) block.}
     \label{figure:IPer_block}
\end{figure}

\noindent\textbf{Invertible Prompter.} We observe that the issues of semantic distortion and information loss could cause irreversible cross-modal generation. To address these issues, we design an invertible prompter, as shown in Figure~\ref{figure:IPer_block}, that introduces an inverse reconstruction scheme in the prompter generation to improve the quality of generated prompts.
To be specific, the invertible prompter block encompasses a bi-directional propagation process.
This design plays a crucial role in establishing a mutual mapping among the available modality, the prompt of the missing modality, and the reconstructed available modality. In forward processes, we ensure the alignment of the prompt with the missing modality. Conversely, in inverse processes, we align the prompt with the input available modality. This strategy enables us to achieve high-quality prompt generation, specifically tailored for the missing modality.
To be more precise, during the forward process of the $i$-th block, we denote the input as $\mathcal{F}^i_{1:2c}$ with a channel dimension of $2c$. First, we split $\mathcal{F}^i_{1:2c}$ evenly into two parts $[\mathcal{F}^i_{1:c},\mathcal{F}^i_{c+1:2c}]$, and then map them to a latent representation using two independent linear neural networks $\mathcal{S}(\cdot)$ and $\mathcal{T}(\cdot)$. The overall process of the $i$-th block can be expressed as follows: 
\begin{equation}\label{eq2}
\begin{aligned}
\mathcal{F}^{i+1}_{1:c} &= \mathcal{F}^i_{1:c} + \mathcal{S}(\mathcal{F}^i_{c+1:2c})\\
\mathcal{F}^{i+1}_{c+1:2c} &= \mathcal{F}^i_{c+1:2c} + \mathcal{T}(\mathcal{F}^{i+1}_{1:c})\\
\mathcal{F}^{i+1}_{1:2c} &= \complement(\mathcal{F}^{i+1}_{1:c}, \mathcal{F}^{i+1}_{c+1:2c}).
\end{aligned}
\end{equation}

In the inverse projection process, we map the prompts of missing modality to the original input available modality using the same parameters. For the $i$-th block, we map $\mathcal{F}^{i+1}_{1:2c}$ to $\mathcal{F}^{i}_{1:2c}$ in a reverse manner. According to Eq.(\ref{eq2}), we can obtain the following equation: 
\begin{equation}\label{eq3}
\begin{aligned}
\mathcal{F}^{i}_{c+1:2c} &= \mathcal{F}^{i+1}_{c+1:2c} - \mathcal{T}(\mathcal{F}^{i+1}_{1:c})\\
\mathcal{F}^{i}_{1:c} &= \mathcal{F}^{i+1}_{1:c} - \mathcal{S}(\mathcal{F}^i_{c+1:2c})\\
\mathcal{F}^{i}_{1:2c} &= \complement(\mathcal{F}^{i}_{1:c}, \mathcal{F}^{i}_{c+1:2c}).
\end{aligned}
\end{equation}
Hence, we follow Eq.(\ref{eq3}) to obtain the original input $\mathcal{F}^{i}_{1:2c}$ from the missing modality prompts $\mathcal{P}^{i+1}_{1:2c}$. As a result, Eqs(\ref{eq2}) and (\ref{eq3}) are inverse functions of each other, and $\mathcal{S}(\cdot)$ and $\mathcal{ T}(\cdot)$ represent continuous linear networks, which ensure that no information is lost during the entire process.

Since data of all modalities are accessible during the training phase, we can introduce a bidirectional mapping loss to guide the learning of invertible prompter, which is defined as follows:
\begin{equation}{\label{eq4}}
\begin{aligned}
Loss_{bm} = \sum_{n= 1}^{N} [(\mathcal{P}_{mm}^n - \mathcal{F}_{mm}^n)^2 + (\hat{\mathcal{P}}^n_{am} - \mathcal{F}_{am}^n)^2].
\end{aligned}
\end{equation}
Here, $\hat{\mathcal{P}}_{am}^n$ represents the output that is reconstructed backward from the prompt features of the missing modality, and is required to simulate the features of available modality. Consequently, through the above loss constraints on the invertible prompter, we can achieve content-preserving cross-modal prompt generation. 

\noindent\textbf{Task Alignment Loss.} To narrow the gap between generated invertible prompts and tracking tasks, we introduce a task alignment loss. It assists in enhancing the discriminability of prompts through two types of losses. The first term, known as prediction alignment loss, aims to minimize the divergence between task-predictive encoding derived from the available modality and prompt features, and those derived from modality-complete scenarios. It can be defined as follows:
\begin{equation}
\begin{aligned} 
Loss_{ta} =  \textit{KL} \left ( \Phi([\mathcal{F}^N_{rgb},\mathcal{P}_{tir}^N]), \Phi([\mathcal{F}^N_{rgb},\mathcal{F}_{tir}^N]) \right ) \\ + \textit{KL} \left ( \Phi([\mathcal{P}^N_{rgb},\mathcal{F}_{tir}^N]), \Phi([\mathcal{F}^N_{rgb}, \mathcal{F}_{tir}^N]) \right )
\end{aligned}
\end{equation}
where $\Phi$ denotes the tracking head network, \textit{KL} refer to the Kullback-Leibler divergence loss~\cite{KL_loss} and $[\cdot]$ indicates channel-wise concatenation. Moreover, we introduce the tracking task loss~\cite{ostrack} for each modality-missing scenarios:
\begin{equation}
\begin{aligned}
Loss_{task} = \ell \left ( \Phi ([\mathcal{P}^N_{rgb},\mathcal{F}_{tir}^N]), B_{gt} \right ) \\ + \ell \left ( \Phi ([\mathcal{F}^N_{rgb},\mathcal{P}_{tir}^N]), B_{gt} \right )
\end{aligned}
\end{equation}
where $\ell$ is primarily composed of classification loss, $\ell_1$ loss, and IoU loss. These losses are applied for target classification and bounding box regression, respectively. $B_{gt}$ represents the ground-truth label. Finally, we implement invertible prompt generation for downstream tracking based on the above two losses.

\subsection{Training \& Inference}
The overall training of the model consists of two stages. In the first stage, we follow the same settings as ~\cite{ostrack} to train the RGBT baseline tracking model $\theta_{M}$. The objective function for the first stage can be expressed as:
\begin{equation}
\underset{{\theta_{M}}}{\min} Loss_{s1} = \ell \left ( \Phi ([\mathcal{F}^N_{rgb},\mathcal{F}_{tir}^N]), B_{gt} \right )
\end{equation}

In the second stage, we freeze the parameters of its backbone and tracking head network. Subsequently, we employ the feature alignment and task alignment losses to optimize all invertible prompters. The objective function for the second stage can be represented as:
\begin{equation}
\begin{aligned}
\underset{{\theta_{IP}}}{\min} Loss_{s2} =  Loss_{task}+ \lambda_a Loss_{ta} + \lambda_b Loss_{bm}
\end{aligned}
\end{equation}
where $\lambda_a$ and $\lambda_b$ are regularization parameters. In our experiments, we set them to 1 and 0.5, respectively.

During inference, we dynamically alter feature extraction paths, depending on the current input state. If RGB modality is missing, the feature extraction paths for TIR modality and the RGB prompt are utilized to predict the target position in the current frame, and vice versa. Further, when the input modality is complete, all invertible prompt paths remain inactive.

\subsection{Difference from Other Prompt Learning Methods}

The proposed approach of invertible prompt learning focuses on integrating the prompt learning mechanism into cross-modal prompt generation to bridge the model gap in modality-missing and modality-complete scenarios. In contrast to existing prompt learning methods in CV or NLP, which focus on adapting between upstream and downstream tasks, IPL also emphasizes the robustness and dynamics of prompt features. 

Previous applications of prompt learning in RGBT tracking~\cite{ViPT,ProRGBTTrack} mainly concentrate on fusing multi-modal features in scenarios with complete modalities. In ProTrack~\cite{ProRGBTTrack}, the concept of prompts is introduced into RGBT tracking by linearly summing the multi-modal images to adapt the tracking model based on RGB modality. In ViPT~\cite{ViPT}, the association between different modalities at various semantic levels is explored, and learning complementary benefits between modalities are established in a prompt learning manner. 

Different from them, IPL introduces the prompt learning strategy into modality-missing RGBT tracking for the first time. Initially, it enhances the model's adaptability to different scenarios with missing modalities by employing the concept of prompt learning. Then, considering the challenges of semantic distortion and information loss during cross-modality prompt feature generation, a novel invertible prompter is designed for content-preserving prompt generation. Therefore, IPL achieves superior performance in the scenarios with missing modalities.

\section{High-quality Benchmark Datasets}
\label{sec:me}

Currently, the impact of modality-missing in RGBT tracking has not been thoroughly investigated. This can be attributed to the fact that existing RGBT tracking datasets can not provide evaluation of modality-missing scenarios. 
To bridge this gap, we construct three high-quality benchmark datasets for comprehensive evaluation of modality-missing RGBT tracking, which are also the first modality-missing RGBT tracking benchmarks. 
To reflect various modality-missing scenarios of real-world challenges, we design the following dataset construction method to construct high-quality modality-missing dataset.

\subsection{Dataset Construction}

\noindent\textbf{Data Collection.} Collecting missing data directly from real-world scenarios is challenging due to the complexity and diversity of missing factors. Consequently, we choose to utilize existing modality-complete datasets as the source data for building the missing datasets. To ensure the diversity of the missing datasets, we select three representative datasets: RGBT234~\cite{li2019rgb234}, the testing set of LasHeR~\cite{li2021lasher}, and the testing set of VTUAV~\cite{Zhang_CVPR22_VTUAV}. These datasets cover a wide range of scenarios, such as surveillance, mobile surveillance, and drones.

\begin{figure}[h]
     \centering
     \begin{tabular}[b]{cc}
     \includegraphics[width=0.8\textwidth]{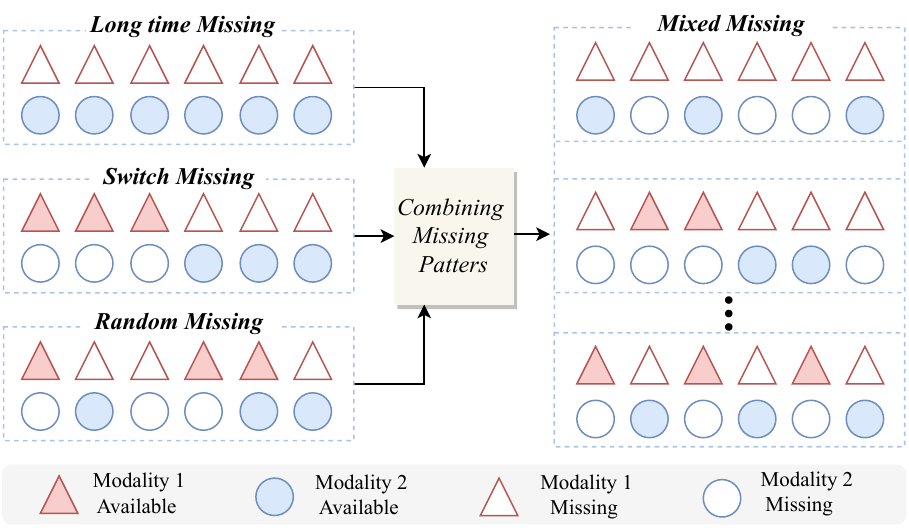}
     \end{tabular}
     \caption{Illustration of different missing patterns and hierarchical combination design.}
     \label{figure:five_miss_pattern}
\end{figure}

\noindent\textbf{Design of Base Missing Patterns.} As shown in Figure~\ref{figure:five_miss_pattern}, we devise three typical missing patterns: long-time missing, switch missing, and random missing. The long-time missing represents continuous missing of one modality, which brings the challenge of transitioning from multi-modality tracking to single-modality tracking. The switch missing denotes the missing alternation between two modalities, which poses a cross-modality tracking challenge~\cite{li2022cross}. Random missing reflects the randomness inherent in real-world scenarios. 

To simulate more complex modality-missing scenarios, we design a hierarchical combination scheme. This scheme combines initial missing conditions with random missing, forming two representative patterns: long-time mixed missing and switch mixed missing. Moreover, we introduce three missing rates (30\%, 60\% and 90\%) to reflect the impact of the number of missing frames. Finally, by combining the five missing patterns with the three missing ratios, we can generate 15 different mixed missing patterns, which cover all real-world modality-missing scenarios.

\noindent\textbf{Dataset Construction.} We construct modality-missing dataset according to the following process. First, we sort all sequences in the given RGBT tracking dataset in descending order based on their frame numbers, and create five groups for the five missing scenarios. Next, we sequentially select the top five sequences and randomly assign them to the corresponding groups until all sequences in the dataset are assigned. Subsequently, we create three subgroups for each group to correspond to the three ratio rates, and assign the sequences of each group to the corresponding three subgroups according to the above process. Finally, we apply the corresponding missing patterns and ratios to each sequence to obtain the modality-missing dataset. In this work, we construct three high-quality modality-missing datasets based on the previously collected datasets, including RGBT234-Miss, LasHeR245-Miss, and VTUAV176-Miss.

\subsection{Dataset Statistics}

\begin{table*}[ht]
\renewcommand\arraystretch{1.7} 
\centering
\caption{Statistics comparison among existing RGBT tracking datasets.}
\resizebox{1\columnwidth}{!}{
\begin{tabular}{ccccccccccc}
\hline
Benchmark  &\begin{tabular}[c]{@{}c@{}} Imaging\\ Platform \end{tabular} & \begin{tabular}[c]{@{}c@{}}Num.\\ Seq.\end{tabular} & \begin{tabular}[c]{@{}c@{}}Total\\ Frame\end{tabular} & \begin{tabular}[c]{@{}c@{}}Avg.\\ Frame\end{tabular} & \begin{tabular}[c]{@{}c@{}}Max.\\ Frame\end{tabular} & \begin{tabular}[c]{@{}c@{}}Total\\ Miss\\ Frame\end{tabular} & \begin{tabular}[c]{@{}c@{}}Avg.\\ Miss\\ Frame\end{tabular} & \begin{tabular}[c]{@{}c@{}}Max.\\ Miss\\ Frame\end{tabular} & Resolution   & Year \\ \hline
GTOT~\cite{li2016gtot}      & Surveillance    & 50                                                  & 7.8K                                                  & 157                                                  & 376                                                  & 0                                                            & 0                                                           & 0                                                           & 384 $\times$  288      & 2016 \\
RGBT210~\cite{Li17rgbt210}   & Surveillance       & 210                                                 & 104.7K                                                & 498                                                  & 4140                                                 & 0                                                            & 0                                                           & 0                                                           & 630 $\times$  460                                                 & 2017 \\ 
RGBT234~\cite{li2019rgb234}  & Surveillance        & 234                                                 & 116.7K                                                & 498                                                  & 4140                                                 & 0                                                            & 0                                                           & 0                                                           & 630 $\times$  460                                                 & 2019 \\
LasHeR~\cite{li2021lasher}   & Surveillance and Mobile        & 1224                                                & 734.8K                                                & 600                                                  & 12862                                                & 0                                                            & 0                                                           & 0                                                           & 630 $\times$  480                                & 2021 \\
VTUAV~\cite{Zhang_CVPR22_VTUAV}    & Drone        & 500                                                 & 1.7M                                                  & 3329                                                 & 27213                                                & 0                                                            & 0                                                           & 0                                                           & 1920 $\times$  1080                               & 2022 \\ \hline
 RGBT234-Miss  & Surveillance   & 234                                                 & 116.7K                                                & 498                                                  & 4140                                                 & 69.2K                                                        & 296                                                         & 1242                                                            & 630 $\times$  460                                                 & 2024 \\
LasHeR245-Miss & Surveillance and Mobile & 245                                                 & 220.7K                                                & 901                                                  & 12862                                                     & 133.2K                                                       & 544                                                         &  3858                                                           & 630 $\times$  480                                                 & 2024 \\
VTUAV176-Miss & Drone  & 176                                                 & 631.5K                                                & 3588                                                 & 25295                                                      & 368.5K                                                       & 2094                                                        & 4097                                                            & 1920 $\times$  1080                                               & 2024 \\ \hline 
\end{tabular}
\label{tab:dataset_compare}
}
\end{table*}

\begin{figure}[h]
     \centering
     \begin{tabular}[b]{cc}
     \includegraphics[width=1\textwidth]{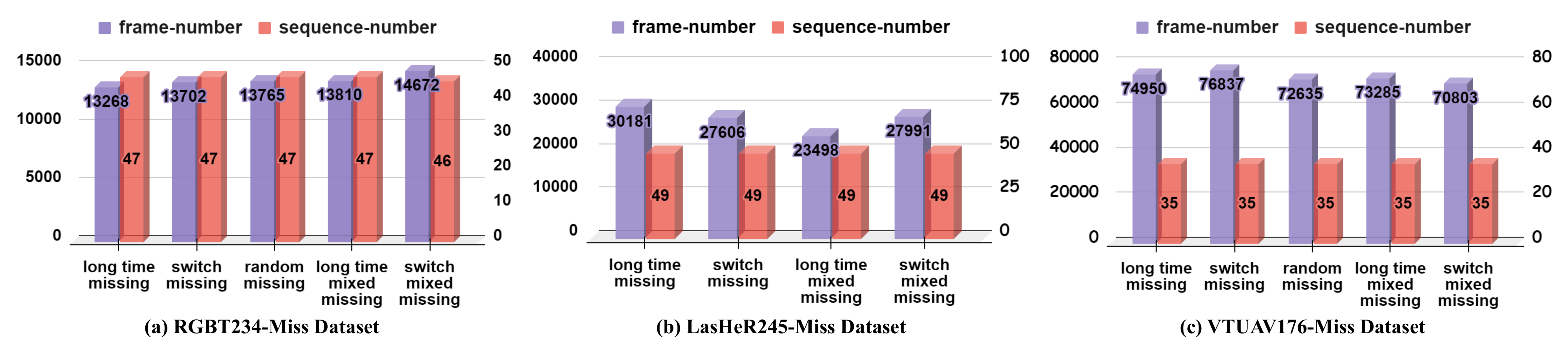}
     \end{tabular}
     \caption{Distribution of the missing patterns in three datasets. The horizontal axis represents the five types of missing patterns, while the left and the right vertical axis represent the number of missing frames and the number of missing sequences, respectively.} 
     \label{figure:modality_miss_patters}
\end{figure}

\noindent\textbf{Attribute Statistics.} We compare the primary attribute information between the generated modality-missing datasets and existing RGBT datasets in Table~\ref{tab:dataset_compare}. We observe that the three modality-missing datasets encompass diverse imaging platform data, thereby satisfying the evaluation requirements of existing trackers. Considering that the challenge of a modality-missing dataset is reflected in the diversity of missing patterns and missing ratios, we analyze the distribution of missing patterns and missing ratios in the three datasets respectively.  Figure~\ref{figure:modality_miss_patters} demonstrates a well-balanced distribution of sequence numbers and frame numbers across different missing patterns. This balanced distribution allows for a more accurate evaluation of tracker performance in specific missing patterns.

\begin{figure}[h]
     \centering
     \begin{tabular}[b]{cc}
     \includegraphics[scale=0.11]{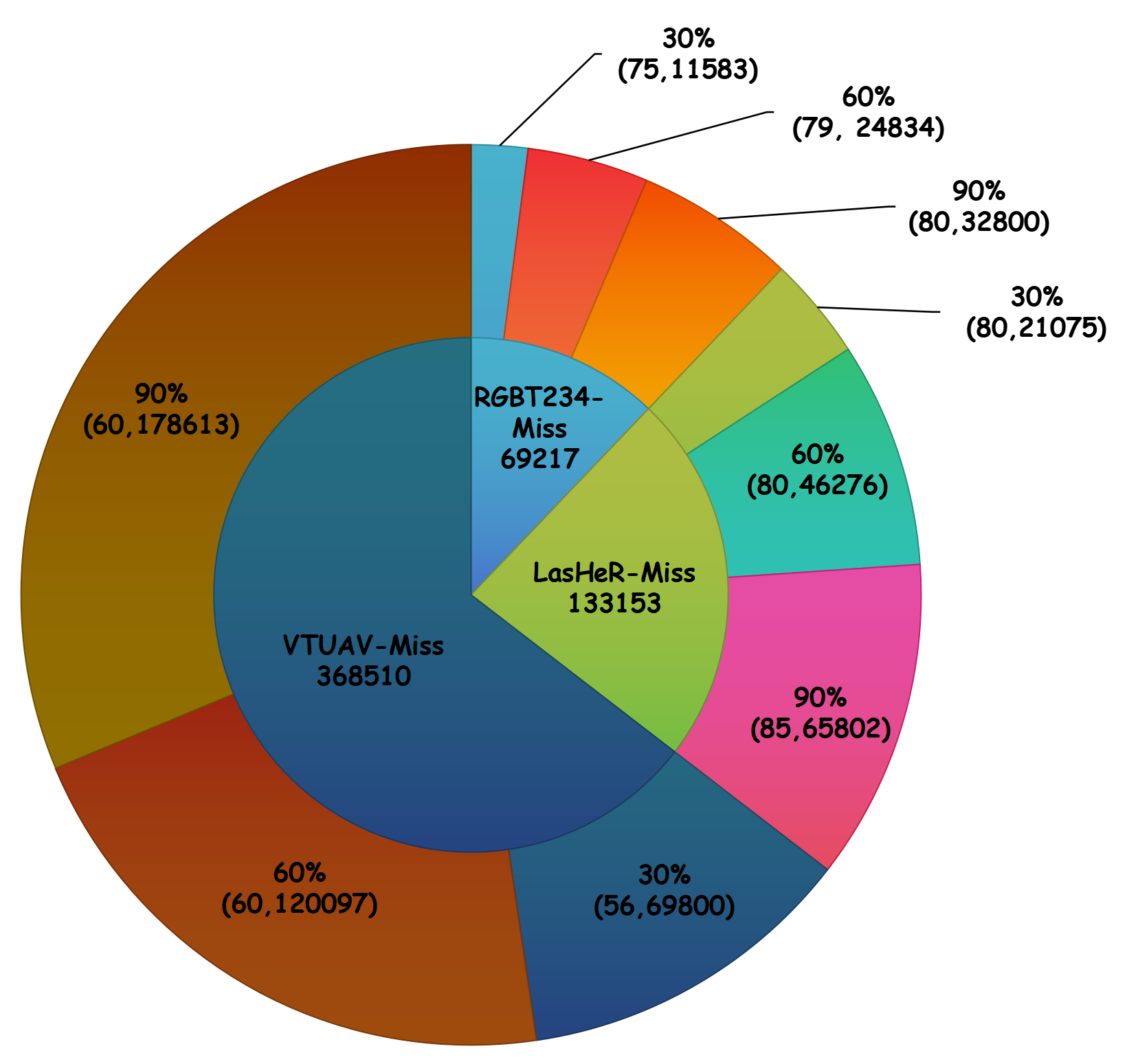}
     \end{tabular}
     \caption{Distribution of the missing ratios in three datasets. The inner circle represents the total number of missing frames in the three datasets, and the outer circle represents the number of missing sequences and missing frames under three missing ratios.}
     \label{figure:modality_miss_ratios_dis}  
\end{figure}

In addition, Figure~\ref{figure:modality_miss_ratios_dis} shows that the number of sequences is well balanced under different missing ratios, while the number of missing frames increases with the increase in the missing ratios. It should be emphasized that we provide annotations of missing patterns and missing ratios for each sequence in the three modality-missing datasets, which helps subsequent research to evaluate RGBT trackers from different perspectives.

\noindent\textbf{Dataset Evaluation.} In our experiments, all trackers follow the One Shot Evaluation (OPE) protocol and evaluate model performance using Maximum Success Rate (MSR) and Maximum Precision Rate (MPR), which are widely used metrics in RGBT tracking~\cite{li2019rgb234, li2021lasher, Zhang_CVPR22_VTUAV}. To fully leverage the capabilities of existing trackers in missing scenes, we provide three data compensation strategies.

\section{Experiments}
\label{sec:ex}

\subsection{Implementation Details}

We implement our model in PyTorch~\cite{paszke2019pytorch} and perform experiments on 6 NVIDIA RTX4090 GPUs. The total batch size for training comprises 64 image pairs. Our model is trained on the LasHeR training set for 60 epochs, each with 60k image pairs. Subsequently, we evaluate the model on RGBT234, LasHeR testing set, RGBT234-Miss and LasHeR245-Miss. For the VTUAV and VTUAV176-Miss dataset. We exclusively use the VTUAV training set for training, while keeping the other settings unchanged. The learning rate of the backbone network is set to 4e-5, and the remaining parameters are set to 4e-4 with a decay factor of 10 every 10 epochs. We utilize AdamW~\cite{AdamW} as the optimizer with a weight decay of 1e-4. The search and template size are adjusted to 256×256 and 128×128, respectively. In the ViT backbone, we introduce modality-specific layers and global modality-specific tokens into layers 4, 7, and 10, while an invertible prompter is embedded in all layers.

\subsection{Comparison with State-of-the-art Methods}

\begin{table*}[]
    \centering
    \caption{Comparison with state-of-the-art methods on RGBT234 and LasHeR testing set for modality-complete and modality-missing scenarios. The best, second and third results are in \textcolor{red}{red}, \textcolor{blue}{blue} and \textcolor{cyan}{cyan} colors, respectively.}
    \label{tab:compare_sota}
    \renewcommand\arraystretch{2}
    \resizebox{1\columnwidth}{!}
    {\begin{tabular}[c]
            {@{}c|c|c|c|c|c|c|c|c|c@{}}
            \toprule
            \multirow{3}{*}{Method}  & \multirow{3}{*}{Pub.info} & RGBT234~\cite{li2019rgb234} & LasHeR~\cite{li2021lasher} & \multicolumn{3}{c}{RGBT234-Miss} & \multicolumn{3}{|c}{LasHeR245-Miss} \\ \cmidrule{3-10}
            & &  \multirow{2}{*}{PR/SR} & \multirow{2}{*}{PR/NPR/SR} & Copy & GAN & Zero & Copy   & GAN & Zero \\ \cmidrule{5-10}
            &&&& PR/SR & PR/SR & PR/SR & PR/NPR/SR & PR/NPR/SR & PR/NPR/SR\\
            \midrule
            
            MANet~\cite{li2019manet}     & ICCVW2019   & 78.4/54.7  & 45.5/38.3/32.6  & 65.7/45.5 & 63.7/43.8 & 63.5/44.0 & 36.6/31.1/31.1 & 34.2/29.1/27.0 & 34.6/29.6/31.1 \\
            mfDiMP~\cite{zhang2019multi} & ICCVW2019   & 82.4/58.3  & 58.3/54.2/45.6  & 69.5/47.3 & 65.6/45.2 & 69.4/48.1 & 37.7/31.8/29.1 & 36.2/30.6/28.0 & 37.0/31.5/27.7 \\
            CAT~\cite{2020CAT}           & ECCV2020    & 80.4/56.1  & 45.0/39.5/31.4  & 52.1/35.6 & 57.5/39.6 & 54.5/37.0 & 28.5/23.8/20.6 & 30.4/25.2/22.5 & 27.0/32.2/20.2 \\
            ADRNet~\cite{ADRNet2021}     & IJCV2021    & 82.6/59.6  & 47.2/42.3/36.1  & 61.0/42.4 & 60.9/42.7 & 60.2/41.8 & 29.4/25.6/21.7 & 31.1/26.8/22.8 & 28.8/24.9/21.3 \\
            DMCNet~\cite{DMCNet2022}    & TNNLS2022    & \textcolor{cyan}{83.9}/59.3  & 49.0/43.1/35.5 & 66.5/45.2 & 61.5/42.0 & 63.3/43.1 & 32.7/28.2/25.3 & 35.2/30.1/26.9 & 33.5/28.7/25.8 \\
            APFNet~\cite{APFNet2022}   & AAAI2022      & 82.7/57.9  & 50.0/43.9/36.2  & 65.3/44.7 & 68.4/47.0 & 68.3/46.6 & 38.2/32.0/29.0 & 39.0/33.0/29.3 & 38.5/32.4/28.8 \\
            ViPT~\cite{ViPT}          & CVPR2023       & 83.5/\textcolor{cyan}{61.7} &\textcolor{cyan}{65.1}/\textcolor{cyan}{61.6}/\textcolor{cyan}{52.5} & 61.2/44.8 & 61.3/44.1 & 52.4/39.3 & 44.0/40.7/36.9 & 47.4/43.7/39.4 & 39.8/37.2/33.8 \\
            TBSI~\cite{TBSI}          & CVPR2023       & \textcolor{blue}{87.1}/\textcolor{blue}{63.7}  & \textcolor{blue}{69.2}/\textcolor{red}{65.7}/\textcolor{red}{55.6} & \textcolor{blue}{76.1}/\textcolor{blue}{55.2} & 73.7/53.0 & \textcolor{cyan}{75.7}/\textcolor{cyan}{54.6} & \textcolor{blue}{59.0}/\textcolor{blue}{54.2}/\textcolor{blue}{47.0} & 56.6/51.8/45.4 & \textcolor{cyan}{57.7}/\textcolor{cyan}{53.4}/\textcolor{cyan}{46.8} \\ 
            \midrule 
            IPL                       & --             & \textcolor{red}{88.3}/\textcolor{red}{65.7}    & \textcolor{red}{69.4}/\textcolor{blue}{65.6}/\textcolor{blue}{55.3} & \multicolumn{3}{c}{\textcolor{red}{82.0}/\textcolor{red}{59.4}} & \multicolumn{3}{|c}{\ \ \ \textcolor{red}{61.7}/\textcolor{red}{56.8}/\textcolor{red}{49.4}} \\

            \bottomrule
    \end{tabular}
}
\end{table*}

\noindent\textbf{Evaluation on Modality-complete Datasets.} We compare our method, denoted as IPL, with the current state-of-the-art RGBT trackers on two modality-complete datasets (RGBT234 and LasHeR). As shown in Table~\ref{tab:compare_sota}, IPL achieves highest or comparable performance in modality-complete scenarios. Specifically, on RGBT234 dataset, IPL outperforms TBSI, the second-best method, by 1.2\%/2.0\% in PR/SR metrics. It also exceeds ViPT and DMCNet by 4.8\%/4.0\% and 4.4\%/6.4\% in PR/SR metrics, respectively. On LasHeR dataset, IPL slightly surpasses TBSI by 0.2\% in PR, while trailing by 0.1\%/0.3\% in NPR/SR metrics. This minor performance disadvantage stems from the fact that TBSI employs highly complex fusion interaction modules, whereas IPL accomplishes multimodal fusion using only simple concatenation operations. Despite this, IPL significantly outperforms ViPT and APFNet in PR/NPR/SR indicators, by 4.3\%/4.5\%/2.8\% and 19.4\%/21.7\%/19.1\%, respectively. These results validate the effectiveness of the proposed method and highlight the advantages of a sharing-specific architecture in modality-complete scenarios.

\noindent\textbf{Evaluation on Modality-missing Datasets.} We also compare IPL with current state-of-the-art RGBT trackers on two modality-missing datasets. As depicted in Table~\ref{tab:compare_sota}, the performance of IPL in the modality-missing scenarios surpasses all advanced trackers that incorporate three missing data compensation strategies: “Copy”, “GAN”, and “Zero”. Specifically, in RGBT234-Miss dataset, IPL outperforms TBSI w/ “Copy” (the second-best method) on the PR/SR metrics by 5.9\%/4.2\%. Furthermore, IPL exceeds the performance of other methods in the PR/NPR/SR metrics in LasHeR245-Miss dataset. For instance, IPL outperforms TBSI w/ “Copy” (the second-best method) and TBSI w/ “Zero” (the third-best method) on the PR/NPR/SR metrics by 2.7\%/2.6\%/2.4\% and 4.0\%/3.4\%/2.6\%, respectively. These results highlight the advantages of the proposed method in modality-missing scenarios. Additionally, it can be observed that the CNN-based tracker combined with the “GAN” compensation strategy performs better in missing scenarios, while Transformer-based tracker performs best with the “Copy” compensation strategy. This implies that different models are suitable for different data compensation strategies.

\noindent\textbf{Discussion of Existing Trackers in Missing Scenarios.} We observe two interesting phenomenons. Firstly, MANet, an RGBT tracking algorithm based on a shared-specific architecture, exhibits moderate performance in modality-complete scenarios. However, in modality-missing scenarios, MANet surpasses several advanced algorithms such as ViPT and ADRNet. This once again highlights the robustness of this architecture in dealing with modality-missing challenges. Secondly, we note that TBSI, due to its unique design approach for the initial template frame, achieves superior performance compared to other advanced algorithms in both modality-complete and modality-missing scenarios. This indicates the importance of the initial template frame information in enhancing performance in modality-missing scenarios. Lastly, by integrating the proposed invertible prompter into TBSI, we further boost its performance in modality-missing scenarios, which suggests the broad applicability of our method.

\begin{table*}[ht]
	\centering
	\caption{Comparison with state-of-the-art methods on VTUAV and VTUAV176-Miss datasets. The best and second results are in \textcolor{red}{red} and \textcolor{blue}{blue} colors, respectively.}
	\label{tab:compare_sota_vtuav}
	\renewcommand\arraystretch{1.2}
	\resizebox{0.65\columnwidth}{!}{
		\begin{tabular}{c|cc|cc|cc}
			\hline
			 \multirow{3}{*}{Method} & \multicolumn{2}{c}{VTUAV~\cite{Zhang_CVPR22_VTUAV}} & \multicolumn{4}{|c}{VTUAV176-Miss}  \\ \cmidrule{2-7}
			& \multirow{2}{*}{PR} & \multirow{2}{*}{SR} & \multicolumn{2}{c}{Copy} & \multicolumn{2}{|c}{Zero} \\ 	\cmidrule{4-7}
			&&&PR&SR & PR&SR \\
			\hline
			HMFT~\cite{Zhang_CVPR22_VTUAV} & 75.8 & 62.7 & 60.5 & 49.2 & 46.4 & 39.1 \\
			mfDiMP~\cite{zhang2019multi}   & 69.4 & 57.1 & 63.2 & 51.8 & \textcolor{blue}{60.0} & 48.0 \\
			ViPT~\cite{ViPT}               & 85.0 & 73.0 & 63.0 & 54.5 & 56.7 & \textcolor{blue}{49.4} \\
                OSTrack$_{rgbt}$~\cite{ostrack}               & \textcolor{blue}{85.0} & \textcolor{blue}{73.4} & \textcolor{blue}{63.3} & \textcolor{blue}{54.5} & 51.8 & 46.1 \\
			\hline

			IPL                            & \textcolor{red}{87.5} & \textcolor{red}{75.6} & & \multicolumn{2}{c}{\textcolor{red}{80.9} \ \ \  \textcolor{red}{68.5}} & \\ 
                \hline
            \end{tabular}}
\end{table*}

\noindent\textbf{Evaluation on VTUAV and VTUAV176-Miss Dataset.} VTUAV is a recently proposed RGBT tracking dataset that focuses on visual tracking in drones. Currently, evaluations of existing algorithms on this dataset are limited. Therefore, we reproduce several advanced RGBT trackers~\cite{zhang2019multi, ViPT, ostrack} on this dataset and compare them with our IPL method. The experimental results are summarized in Table~\ref{tab:compare_sota_vtuav}.
In modality-complete scenarios (VTUAV dataset), IPL performs excellently in both PR and SR metrics. Specifically, compared to the second-best method, OSTrack$_{rgbt}$, IPL improves by 2.5\% and 2.2\% on these two metrics respectively, demonstrating a significant advantage.
In modality-missing scenarios, on VTUAV176-Miss dataset, IPL obtains 80.9\% and 68.5\% on the PR and SR metrics respectively, both of which are the best among all methods. Specifically, IPL outperforms OSTrack$_{rgbt}$ w/ “Copy” (the second-best method) by 17.6\% and 14.0\% on the PR and SR metrics respectively. 
These results clearly demonstrate that IPL surpasses current state-of-the-art methods in both modality-complete and modality-missing scenarios. These results further validate the effectiveness and significant advantages of our proposed method in drone scenario tracking.


\noindent\textbf{Evaluation on Different Modality-missing Challenges.} In modality-missing scenarios, we establish two types of challenges, including various modality missing patterns and different modality missing ratios. Consequently, we conduct an in-depth evaluation of our method on different missing challenge subsets within three modality-missing datasets (RGBT234-Miss, LasHeR245-Miss, and VTUAV176-Miss).

\begin{figure}[ht]
     \centering
     \begin{tabular}[b]{cc}
     \includegraphics[width=1\textwidth]{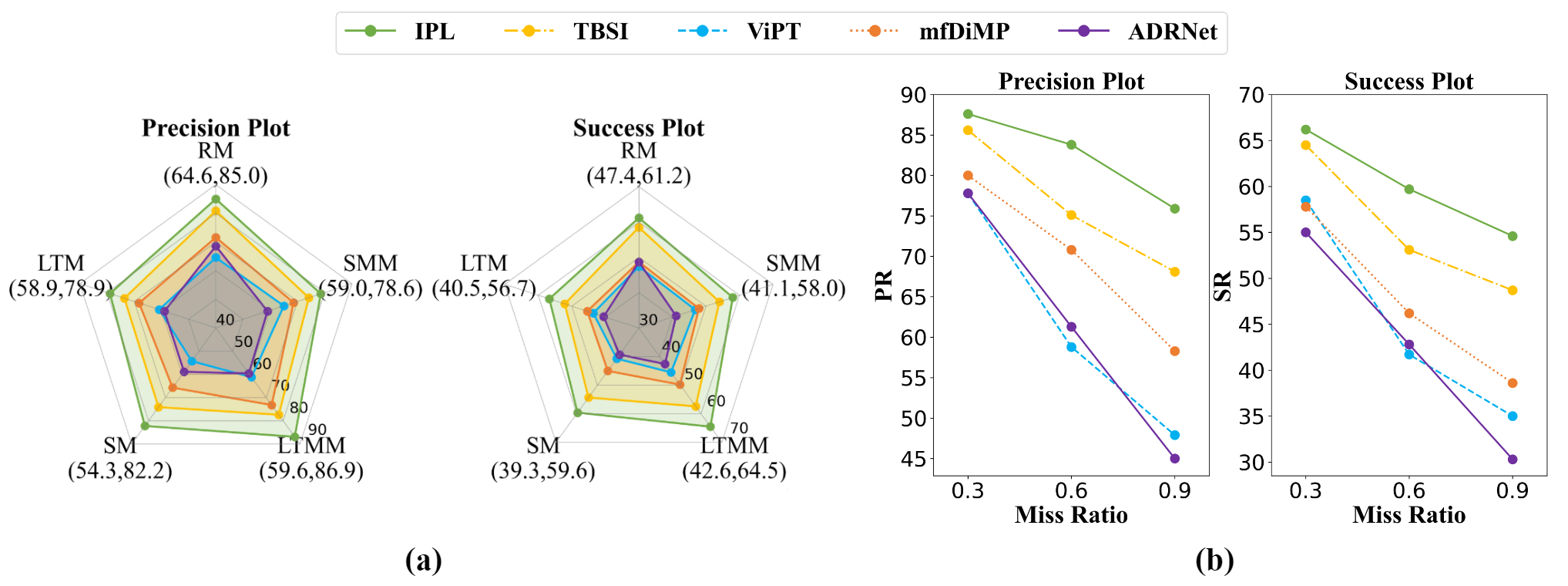}
     \end{tabular}
     \caption{Comparison of the proposed method with state-of-the-art RGBT trackers on different missing pattern (a) and miss ratio (b) subsets of RGBT234-Miss dataset.}
     \label{figure:modality_miss_patters_exp}
\end{figure}

\begin{figure}[ht]
     \centering
     \begin{tabular}[b]{cc}
     \includegraphics[scale=0.09]{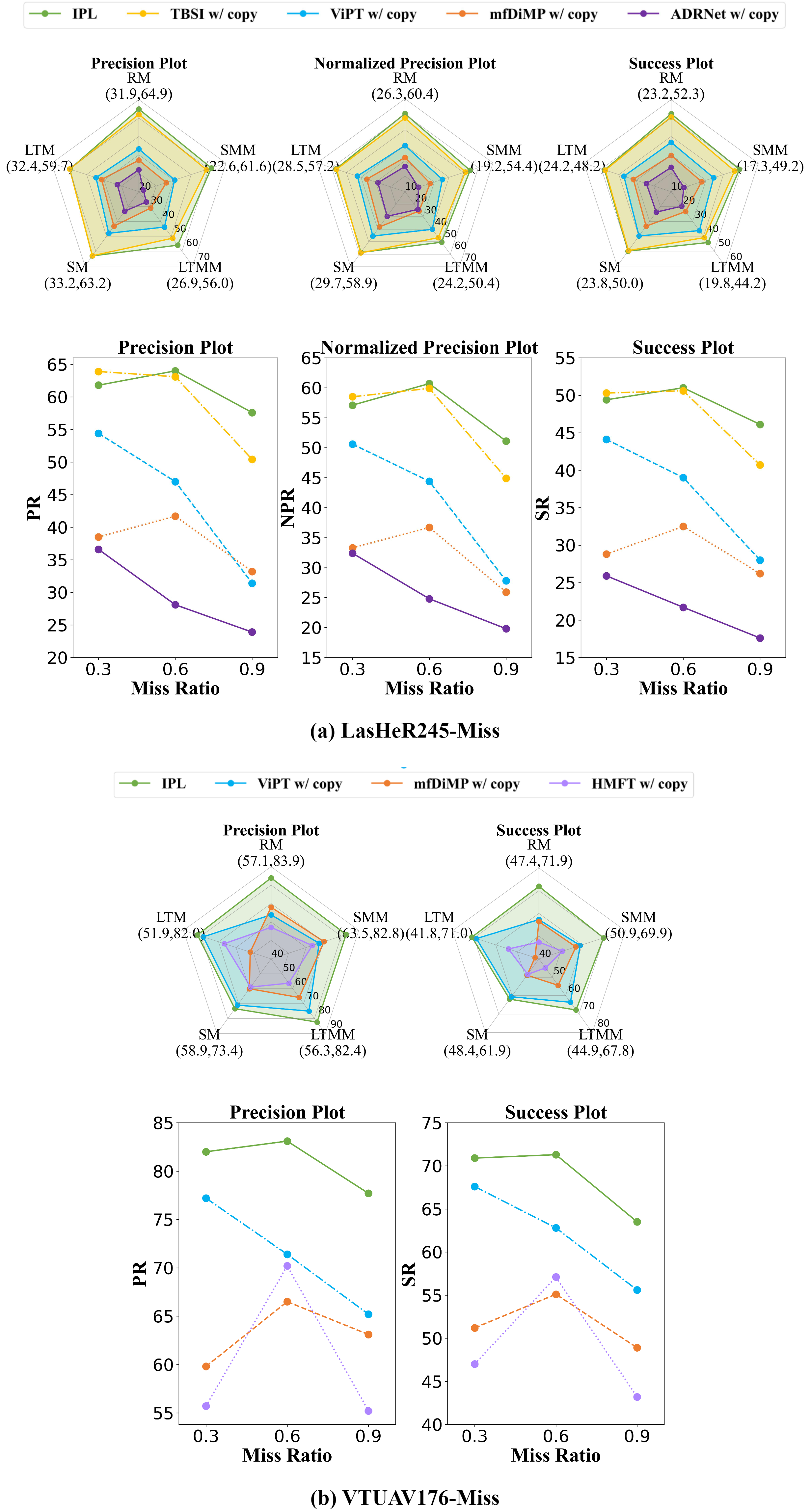}
     \end{tabular}
     \caption{Comparison of IPL with state-of-the-art RGBT trackers on different missing pattern subsets.}
     \label{figure:modality_miss_patterns_vtuav}  
\end{figure}

On RGBT234-Miss dataset, we analyze the performance of our method across different missing pattern subsets, including Long Time Missing (LTM), Switch Missing (SM), Random Missing (RM), Long Time Mixed Missing (LTMM), and Switch Mixed Missing (SMM). As shown in Figure~\ref{figure:modality_miss_patters_exp} (a), our method surpasses previous state-of-the-art trackers in all subsets. Notably, the performance degradation is more significant in the LTM, LTMM, and SMM subsets on both datasets, indicating that different missing pattern designs pose distinct challenges. In addition, we also analyze the performance of our method across different missing ratio subsets, including 30\%, 60\%, and 90\%. As shown in Figure~\ref{figure:modality_miss_patters_exp} (b), the performance of all methods declines as the missing ratio increases. However, our method consistently displays strong robustness and excellent performance.

On LasHeR245-Miss and VTUAV176-Miss datasets, we conduct performance comparison experiments in the missing pattern subsets and missing ratio subsets. As depicted in Figure~\ref{figure:modality_miss_patterns_vtuav} (a), our method achieves a leading performance advantage in most missing pattern and missing ratio subsets compared to existing advanced methods. However, when the missing ratio is 30\%, the performance of our method is marginally lower than TBSI method. This can be attributed to the limited length of continuous missing frames when the missing ratio is low, and the target template in TBSI can play a significant role. However, as the missing ratio increases, this effect gradually weakens. In Figure~\ref{figure:modality_miss_patterns_vtuav} (b), it can be observed that our method outperforms previous state-of-the-art trackers in all missing pattern and missing ratio subsets on VTUAV176-Miss. Furthermore, the performance degradation is more pronounced in SM subset of VTUAV176-Miss dataset and LTMM subset of LasHeR245-Miss. This also shows again that different missing patterns pose different challenges.

\subsection{Ablation Study}
 
\noindent{\textbf{Component Analysis under Modality-missing Datasets.}} Table \ref{tab:ablation_exp1} shows the ablation results of our IPL method in modality-missing scenarios. We conduct a comprehensive analysis of each component of IPL by comparing its PR and SR performance on RGBT234-Miss and LasHeR245-Miss datasets. In this experiment, we denote our proposed method as “Base w/ IPer(deep) TA” and gradually remove key components to evaluate their impacts. Firstly, we examine the effect of reducing the number of IPer modules by embedding IPer only in the first layer while keeping other components unchanged, referred to as “Base w/ IPer(shallow) TA”. The decrease in model performance with fewer IPer modules indicates the effectiveness of multi-level IPer. Moreover, “Base w/ IPer(shallow)TA” still outperforms our baseline method (“Base w/ copy”), suggesting that the shallow prompt generation strategy is also effective.

\begin{table}[ht]
\renewcommand\arraystretch{1.4}
\caption{Ablation analysis of IPL in modality-missing datasets.}
\label{tab:ablation_exp1}
\begin{tabular*}{\textwidth}{@{\extracolsep\fill}l|ccccc}
       \hline
      \multirow{2}{*}{Case} & \multicolumn{2}{c}{RGBT234-Miss} &  \multicolumn{3}{c}{LasHeR245-Miss} \\ & PR   & SR & PR &NPR  & SR \\
      \hline
        Base w/ IPer(deep) TA & 82.0            & 59.4 & 61.7 & 56.8  & 49.4  \\ 
        Base w/ IPer(shallow) TA & 80.6            & 58.8 & 59.7 & 55.2  & 47.9  \\  
        Base w/ LoRA(deep) TA & 79.1            & 57.4     &59.1 & 54.6 & 47.5     \\
        Base w/ IPer(deep)    & 80.3   & 58.5 & 60.2  & 55.5  & 48.2   \\
        Base w/ Copy  & 74.8   & 54.4   & 57.7 & 53.2  & 46.5 \\
        \hline
\end{tabular*}
\end{table}

Subsequently, in order to verify the importance of the content-keeping prompt generation strategy designed in this paper, we use the currently widely used LoRA module to replace all IPer modules in the model, and keep other components unchanged, denoted as “Base w/ LoRA(deep) TA”. It can be observed from Table~\ref{tab:ablation_exp1} that “Base w/ LoRA(deep) TA” shows significant performance degradation compared with “Base w/ IPer(deep) TA” in both missing datasets. In other words, the PR/SR metric of RGBT234-Miss decreases by 2.9\%/2.0\%, and the PR/NPR/SR metric of LasHeR245-Miss decreases by 2.6\%/2.2\%/1.9\%, respectively. In addition, to verify the validity of the designed task alignment loss (TA), we remove this loss from the model and denote it as “Base w/ IPer(deep)”. It can be observed that compared with “Base w/ IPer(deep) TA”, there is a certain performance degradation in both missing datasets. That is, the decrease in the PR/SR metric of RGBT234-Miss is 1.7\%/0.9\%, and the decrease in the PR/NPR/SR metric of LasHeR245-Miss is 1.5\%/1.3\%/1.2\%.

Finally, we remove all components to get the baseline model, and  adopt the “Copy” data compensation strategy to cope with the modal missing input, denoted as “Base w/ Copy”. It can be observed that our overall approach has significant performance advantages over the baseline model. Specifically, the PR/SR metric of RGBT234-Miss exceeds 8.8\%/5.0\%, and the PR/NPR/SR metric of LasHeR245-Miss exceeds 4.0\%/3.6\%/2.9\%, respectively. In summary, through this series of ablation experiments, we verify the importance of invertible prompters and task alignment losses in IPL method, and also confirm the superiority of multi-layer invertible prompters in deep networks.

\begin{table}[]
\renewcommand\arraystretch{1.4}
\caption{Ablation analysis of IPL in modality-complete datasets.}
\label{tab:ablation_exp2}
\begin{tabular*}{\textwidth}{@{\extracolsep\fill}c|ccccccc}
       \hline
      \multirow{2}{*}{Case} & \multicolumn{2}{c}{RGBT234~\cite{li2019rgb234}} &  \multicolumn{3}{c}{LasHeR~\cite{li2021lasher}} & \multicolumn{2}{c}{VTUAV~\cite{Zhang_CVPR22_VTUAV}}  \\ & PR   & SR   & PR & NPR  & SR & PR   & SR \\
      \hline
      Base  & 86.4            & 64.5  & 65.5   &  62.0      & 52.1 &  85.0      & 73.4 \\
      IPL   & 88.3            & 65.7 & 69.4   &  65.6      & 55.3  &  87.5      & 75.6 \\
        \hline
\end{tabular*}
\end{table}

\noindent{\textbf{Component Analysis under Modality-complete Datasets.}} We also verify the validity of the proposed method in a modality-complete datasets. Since the invertible prompter will not be enabled in modality-complete scenarios, the only difference from the baseline method (Base) is shared-specific architecture. As shown in Table \ref{tab:ablation_exp2}, our method (IPL) shows significant advantages over the baseline method in the three mainstream modality-complete datasets.


\begin{table}[]
\renewcommand\arraystretch{1.4}
\caption{Performance comparison of different trackers with missing and non-missing training algorithms.}
\label{tab:ablation_exp3}
\begin{tabular*}{\textwidth}{@{\extracolsep\fill}l|ccccc}
\hline
\multirow{2}{*}{Case}             & \multicolumn{2}{c}{RGBT234-Miss} & \multicolumn{3}{c}{LasHeR245-Miss} \\
                              & PR              & SR             & PR         & NPR       & SR        \\ \hline
ViPT~\cite{ViPT} w/ copy  &   61.2              &  44.8              &   44.0         & 40.7          & 36.9          \\
ViPT~\cite{ViPT}* w/ copy  &   60.5              &  43.9              &   43.0         & 39.7          & 36.1          \\
TBSI~\cite{TBSI} w/ copy & 76.1            & 55.2           &   59.0         &  54.2         &  47.0         \\
TBSI~\cite{TBSI}* w/ copy & 77.7            & 56.8           &   59.8         &  55.2         &  48.1         \\ \hline
IPL                           & 82.0            & 59.4           & 61.7       & 56.8      & 49.4      \\ \hline
\end{tabular*}
\end{table}

\noindent{\textbf{Analysis of Training with Missing and Non-missing schemes.}} To investigate the influence of introducing input from modal-missing scenarios into the training set, we retrain two state-of-the-art RGBT trackers, ViPT and TBSI. Specifically, during training, we employ the “Copy” data compensation strategy to construct the model input types in missing scenarios, which include: RGB-RGB, TIR-TIR, and RGB-TIR. In Table \ref{tab:ablation_exp3}, TBSI* and ViPT* represent the training methods using modality-missing data, while TBSI and ViPT represent the training methods using no modality-missing data. We observe that the performance of TBSI* in the evaluation of the modality-missing dataset improves, while the performance of ViPT* declines. This indicates that training methods that introduce modality-missing exhibit inconsistent effects on trackers. In addition, IPL achieves superior performance using a similar training strategy compared to the modest improvement TBSI* achieves.

\noindent{\textbf{Other Ablation Experiments.}} Apart from verifying the effectiveness of individual components, we also perform two sets of analysis experiments to demonstrate the significance and value of invertible prompt learning. In Table~\ref{tab:ablation_exp4} (a), we train three trackers based on OSTrack~\cite{ostrack}, including one RGBT tracker, one RGB tracker, and one TIR tracker. Then the corresponding tracker is executed adaptively according to the input type in the modality-missing scenario. It can be seen that the method of training multiple trackers according to all missing scenarios, denoted as “Switch”, can effectively improve the tracking performance in the modality-missing scenario, but it also inevitably brings a large number of parameters. Conversely, IPL achieves superior performance with fewer parameters. In Table \ref{tab:ablation_exp4} (b), it can be observed that embedding the IPL scheme in TBSI~\cite{TBSI}, denoted as “IPL$_{TBSI}$”, can improve its performance in modality-missing scenario, which indicates the generality of IPL.


\begin{figure}[h]
  \captionof{table}{Other ablation analysis of IPL.}
  \label{tab:ablation_exp4}
\renewcommand\arraystretch{1.4}
  \begin{minipage}[t]{0.4\textwidth}
    \centering
    \setlength\tabcolsep{3pt}
    \begin{adjustbox}{valign=t}
       \begin{tabular}{l|ccc}
       \hline
      \multirow{2}{*}{Case} & \multicolumn{2}{c}{RGBT234-Miss} &\multirow{2}{*}{Params}\\ & PR   & SR \\
      \hline
        Base  & 74.8   & 54.4  & 92.1 M \\
        Switch  & 78.9           & 57.7     & 276.3 M \\
        IPL    & 82.0            & 59.4    & 183.6 M \\
        \hline
      \end{tabular}
    \end{adjustbox}
    \subcaption{\textbf{Prompt} vs \textbf{Switch}}
  \end{minipage} \qquad 
\hspace{0.02\textwidth}
  \begin{minipage}[t]{0.5\textwidth}
    \centering
    \setlength\tabcolsep{3pt}
    \begin{adjustbox}{valign=t}
    \begin{tabular}{l|ccccc}
    \hline
      \multirow{2}{*}{Case} & \multicolumn{2}{c}{RGBT234-Miss} & \multicolumn{3}{c}{LasHeR245-Miss} \\ & PR   & SR & PR &NPR  & SR    \\
      \hline
        TBSI   & 76.1            & 55.2    & 59.0     & 54.2        & 47.0  \\
        TBSI* & 77.7            & 56.8           &   59.8         &  55.2         &  48.1         \\
        IPL$_{TBSI}$    & 79.5            & 58.0   & 62.2     & 57.1        & 49.4 \\
        \hline
      \end{tabular}
    \end{adjustbox}
    \subcaption{\textbf{IPL in different trackers}}
  \end{minipage}
\end{figure}

\begin{figure*}[ht]
     \centering
     \begin{tabular}[b]{cc}
     \includegraphics[width=0.8\textwidth]{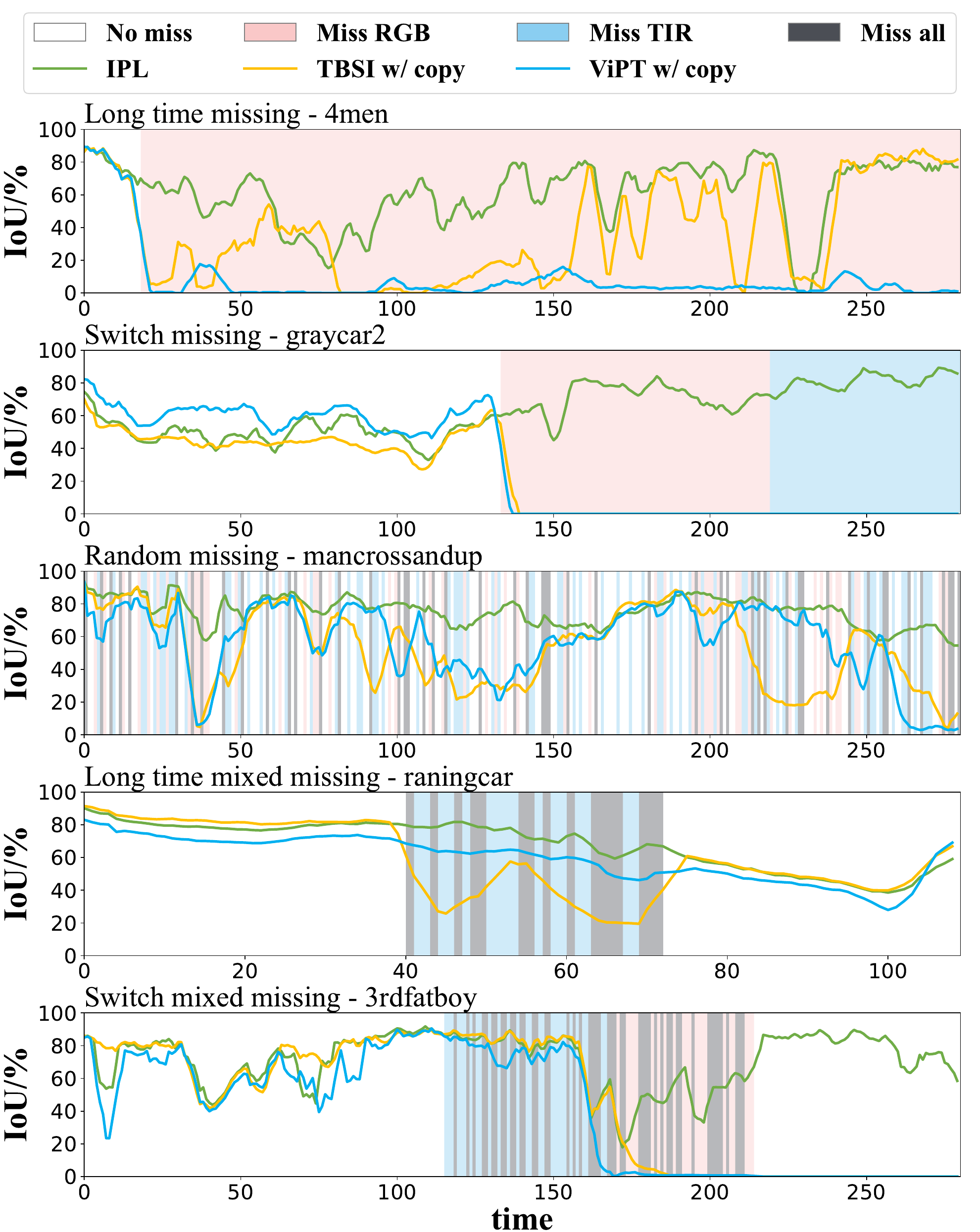}
     \end{tabular}
     \caption{Qualitative comparison between IPL and advanced trackers on five representative sequences.}
     \label{figure:modality_miss_iou}
\end{figure*}

\begin{figure}[ht]
     \centering
     \begin{tabular}[b]{cc}
     \includegraphics[width=1\textwidth]{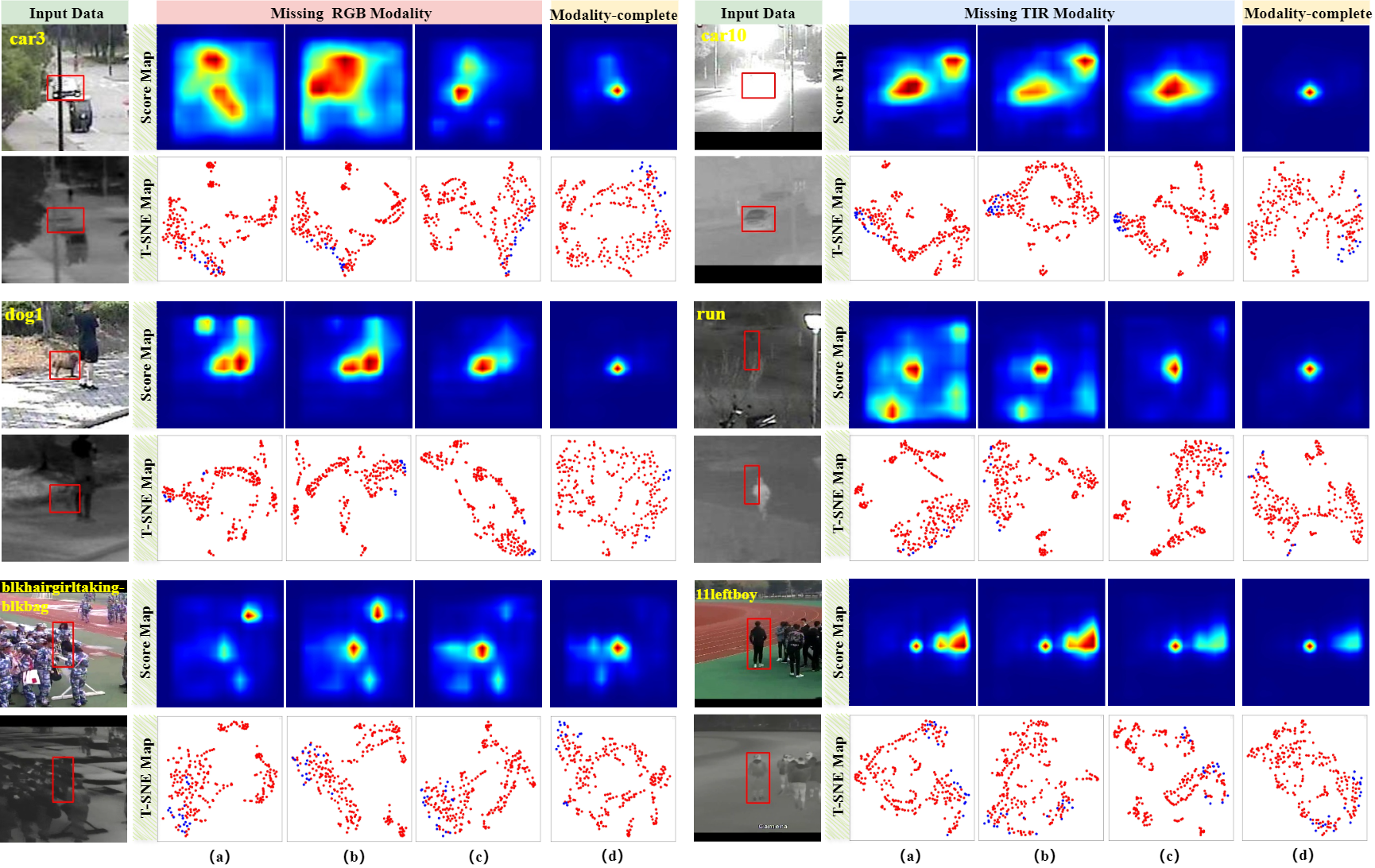}
     \end{tabular}
     \caption{Visualization of score maps and t-SNE maps of different models. The red boxes denote ground truth. (a) OSTrack$_{rgbt}$, (b) IPL w/o IPer, (c) IPL with missing modalities, (d) IPL with complete modalities.}
     \label{figure:feature_vis_map_support}  
\end{figure}

\noindent{\textbf{Qualitative Comparison.}} As shown in Figure~\ref{figure:modality_miss_iou}, we perform a qualitative comparison between our method and two Transformer-based RGBT trackers. Five representative sequences covering five different missing pattern types are used to evaluate the tracking stability of different methods under missing scenarios. For instance, in the \textit{graycar2} sequence, existing trackers fail to maintain tracking when they encounter missing modalities, while our tracker leverages invertible prompt learning to achieve stable tracking performance in modality-missing scenes. In other sequences, our method also effectively addresses various types of challenges associated with missing modalities, including long-term mixed missing, switching mixed missing, and random missing. These results indicate that our invertible prompt learning method is effective in enhancing the robustness of tracking model on different missing patterns, which is also verified in Figure~\ref{figure:modality_miss_patters_exp} and Figure~\ref{figure:modality_miss_patterns_vtuav}.

In Figure~\ref{figure:feature_vis_map_support}, we present the score maps and t-SNE maps of three model outputs: the baseline model OSTrack$_{rgbt}$, the IPL model without an invertible prompter (IPL w/o IPer), and the IPL for both scenarios with missing modalities and complete modalities. In the scenarios with missing modalities, the score map of the baseline model exhibits significant responses in the non-target regions, which leads to model drift. However, the IPL w/o IPer model, benefiting from the shared-specific design architecture, improves the score map compared to the baseline model. Nevertheless, there is still a significant gap when compared to the score map in the scenarios with complete modalities. These phenomenons are also reflected in the t-SNE maps. In IPL, it can be seen that our method accurately provides the response of the target regions in both scenarios with missing modalities and complete modalities. This demonstrates that IPL effectively mitigates the adverse effects of modality-missing challenge.

\section{Conclusion}
\label{sec:con}

In this paper, we propose a novel invertible prompt learning approach to address the issue of missing modalities in RGBT tracking, which effectively improves the performance of existing RGBT trackers in modality-missing scenarios through a content-preserving prompt generation scheme. In addition, we also create three high-quality modality-missing RGBT tracking benchmark datasets, which make three strong contributions. First, we solve the problem of lack of unified benchmarks for modality-missing RGBT tracking. Second, we annotate five missing patterns and three missing ratios at the sequence level, meeting the evaluation requirements of trackers under specific missing challenges. Third, to the best of our knowledge, this is the first study to address the issue of missing modalities in RGBT tracking, which can bring insights to other multi-modality tracking tasks.

\bibliography{sn-bibliography}

\end{document}